\documentclass[10pt,twocolumn,letterpaper]{article}
\usepackage{subcaption} 
\usepackage{multirow}
\usepackage{amsfonts}
\usepackage{amssymb}
\usepackage{amsmath}
\usepackage{authblk}
\usepackage[misc]{ifsym} 
\usepackage[pagenumbers]{iccv} 

\definecolor{iccvblue}{rgb}{0.21,0.49,0.74}
\usepackage[pagebackref,breaklinks,colorlinks,allcolors=iccvblue]{hyperref}
\def\paperID{12921} 
\def\confName{ICCV}
\def\confYear{2025}

\title{Lightweight Adapter Learning for More Generalized Remote Sensing Change Detection}

\author[a]{Dou Quan$^{\textrm{\Letter}}$}
\author[a]{Rufan Zhou}
\author[a]{Shuang Wang}
\author[b]{Ning Huyan}
\author[a]{Dong Zhao}
\author[c]{Yunan Li}
\author[a]{Licheng Jiao}
\affil[ ]{$^{\textrm{\Letter}}$Corresponding Author, \textit{quandou@xidian.edu.cn}}
\affil[a]{School of Artificial Intelligence, Xidian University}
\affil[b]{Department of Automation, Tsinghua University}
\affil[c]{School of Computer Science and Technology, Xidian University}

\begin{document}
\maketitle
\begin{abstract}
Deep learning methods have shown promising performances in remote sensing image change detection (CD). However, existing methods usually train a dataset-specific deep network for each dataset. Due to the significant differences in the data distribution and labeling between various datasets, the trained dataset-specific deep network has poor generalization performances on other datasets. To solve this problem, this paper proposes a change adapter network (CANet) for a more universal and generalized CD. CANet contains dataset-shared and dataset-specific learning modules. The former explores the discriminative features of images, and the latter designs a lightweight adapter model, to deal with the characteristics of different datasets in data distribution and labeling. The lightweight adapter can quickly generalize the deep network for new CD tasks with a small computation cost. Specifically, this paper proposes an interesting change region mask (ICM) in the adapter, which can adaptively focus on interested change objects and decrease the influence of labeling differences in various datasets. Moreover, CANet adopts a unique batch normalization layer for each dataset to deal with data distribution differences. Compared with existing deep learning methods, CANet can achieve satisfactory CD performances on various datasets simultaneously. Experimental results on several public datasets have verified the effectiveness and advantages of the proposed CANet on CD. CANet has a stronger generalization ability, smaller training costs (merely updating 4.1\%-7.7\% parameters), and better performances under limited training datasets than other deep learning methods, which also can be flexibly inserted with existing deep models.
\end{abstract}     
\section{Introduction}
Image change detection aims to detect the interested and changed regions in multi-temporal images \cite{li2023new}, which is widely used in land resource management and land uses, urban management, and disaster information estimation \cite{shi2021deeply, gong2015change}. 

\begin{figure}[t]
\centering
\includegraphics[width=0.8\linewidth]{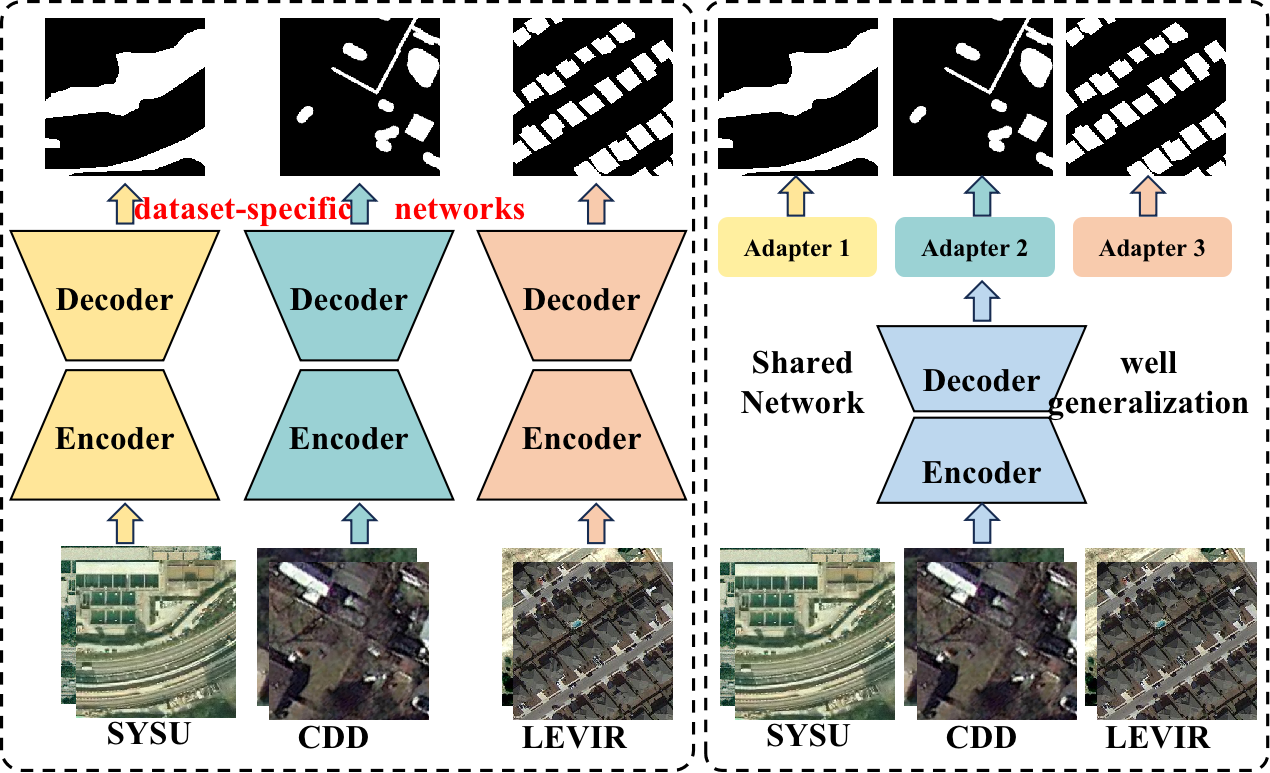}
\caption{Existing deep learning methods usually train a dataset-specific deep network for different datasets, but their generalization performance is poor on other datasets. This paper proposes a change adapter network for more generalized change detection, which can perform well on various datasets simultaneously.} 
\label{introductionfigure}
\vspace{-3mm}
\end{figure}

Recently, due to the powerful feature extraction capabilities, deep learning methods have shown significant performance improvements in image change detection \cite{feng2023change, shao2021sunet}. Deep learning methods usually adopt the encoder-decoder structure for binary change detection label prediction, such as changed and unchanged \cite{daudt2018fully}. The encoder extracts discriminative features from multi-temporal images, and the decoder recovers the resolution of feature maps by upsampling and multi-level feature fusion. However, existing deep learning methods usually train a dataset-specific deep network for each dataset, as shown in Fig. \ref{introductionfigure}. We observed that these dataset-specific deep networks are fully independent and have very poor generalization performances on other datasets. That is to say, the trained deep network based on historical datasets cannot be used for new datasets. This paper analyzes the main reason for the weak generalization performance: significant differences in the data distribution and labeling between various datasets. 


\begin{itemize}
 \item \textbf{ Data distribution differences.} There are different image styles in various datasets. The images may be captured by different sensors or at different times, leading to different data distributions and decreasing the generalization performances of deep networks. 

\item \textbf{Labeling differences.} There are significant labeling differences between datasets caused by various interested objects and different labeling granularities. Firstly, there are different objects of interest in various datasets, such as building objects, roads, vehicles, and so on. Secondly, there are different labeling granularities between datasets, such as detailed change labeling and coarse change labeling. 



\end{itemize}
 
Data distribution and labeling differences between various datasets increase the challenges of deep model generalization. Thus, this paper aims to build a more generalized deep network that can perform CD well on various tasks simultaneously, avoiding training a dataset-specific deep network for each dataset and significantly decreasing memory and training costs. 

To achieve this, this paper proposes a change adapter network (CANet) for more generalized change detection. Considering both the common feature learning goal and significant differences, CANet contains two components: a dataset-shared learning module and a dataset-specific learning module. The dataset-shared learning module leverages a shared convolutional network to extract discriminative features from images. To deal with the labeling differences of various datasets and adapt to dataset characteristics, CANet designs a lightweight dataset-specific learning module, adapter. Specifically, CANet designs an interesting change region mask (ICM) in the adapter module to adaptively guide the interested objects CD. Additionally, CANet adopts a unique batch normalization (BN) \cite{ioffe2015batch} layer for each dataset to decrease the influence of distribution differences. Unlike existing deep learning networks merely applied for a single dataset, the proposed CANet has strong generalization performances and can achieve better results for multiple datasets simultaneously. Moreover, CANet reuses network parameters as much as possible, and most of the network parameters are shared between datasets. Thus, CANet can be updated and adapted for new datasets by updating a few parameters in the adapter, improving parameter efficiency, computation efficiency, and generalization.

The main contributions of this paper are as follows:
\begin{enumerate}
 \item This paper presents the insight analysis of existing deep learning CD methods with poor generalization performances on various datasets caused by excellent differences in the data distribution and labeling. They usually train a dataset-specific deep network for each dataset, significantly increasing the parameter storage and computation cost. 

\item This paper proposes a change adapter network (CANet) for more generalized CD. CANet designs a lightweight dataset-specific learning module, adapter, to adapt different dataset characteristics. CANet proposes an ICM in the adapter module and adopts unique BN layers to deal with the differences in data distribution and labeling, respectively. 

\item Extensive experiments have shown the advantages and effectiveness of the proposed CANet, which has strong generalization performances and achieves better CD results on various datasets simultaneously. CANet can be quickly applied to new datasets by updating a few parameters, decreasing the training cost and risk of overfitting under limited labeled datasets.  
\end{enumerate}

\section{Related work}

Image change detection is used to pinpoint areas of difference within multi-temporal images taken at different times that have been spatially aligned, which is crucial for remote sensing image analysis and applications \cite{khelifi2020deep, shi2020change}. The change detection methods can be divided into traditional and deep learning approaches.
\subsection{Traditional Method}
In the early days, constrained by the scarcity and lower resolution of available remote sensing images, the focus on change objects primarily centered around buildings, roads with a limited number of classes. In this context, traditional change detection methods could achieve relatively robust performance. Traditional change detection methods are mainly classified into algebra-based and transform-based methods. Algebra-based change detection method mainly obtains the change map through simple mathematical operations \cite{singh1989review}, such as image difference \cite{bruzzone2000automatic}, image regression \cite{ridd1998comparison}, etc. Some transform-based methods enhance the change information by transforming the remote sensing images into other spaces \cite{zhong2007multiple}, including principal component analysis (PCA) \cite{celik2009unsupervised} and tasseled cap transformation \cite{crist1985tm}. Traditional change detection methods are sensitive to lighting changes, noise effects, and image style changes, resulting in many pseudo changes.

\begin{figure*}[!t]
\centering
\includegraphics[width=1\linewidth]{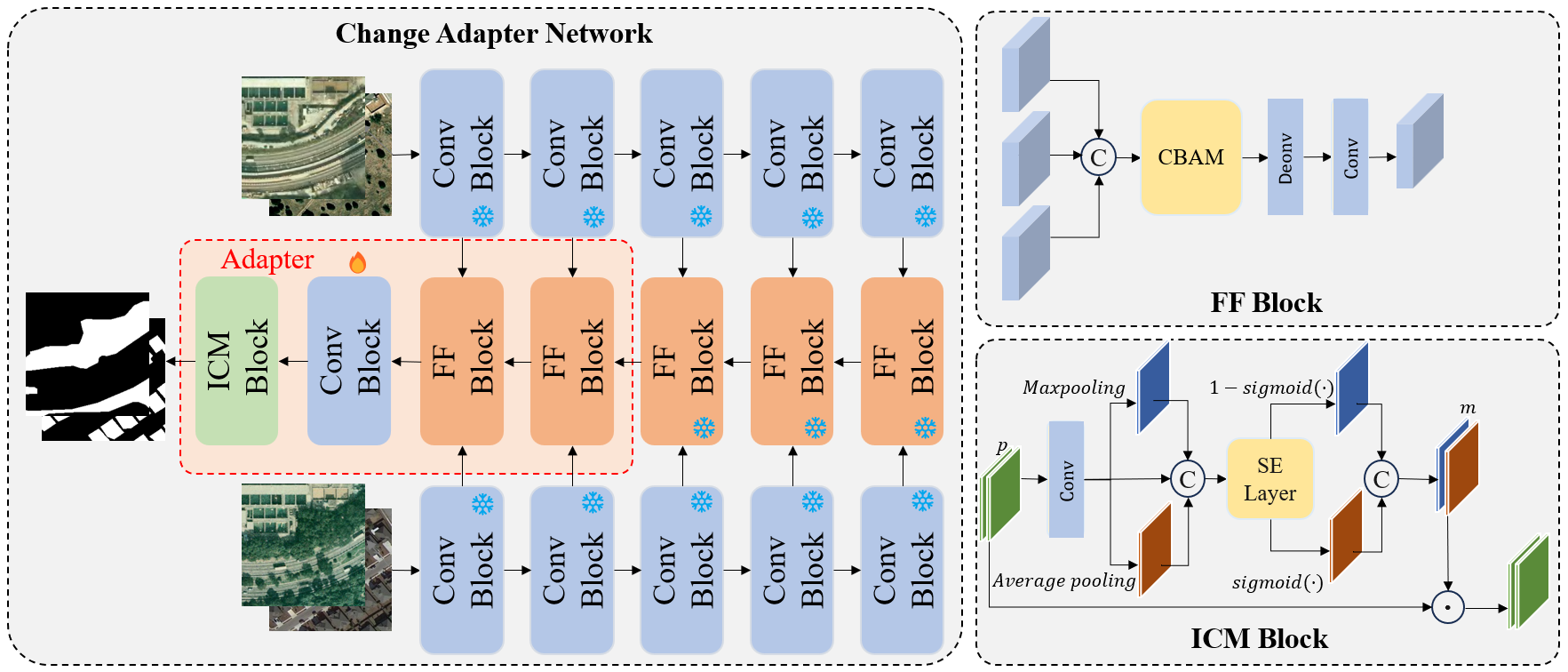}
\caption{Framework of the proposed change adapter network, CANet, for change detection. CANet contains a dataset-shared learning module and a dataset-specific learning module, adapter.}
\label{FRAMEWORK}
\end{figure*}

\subsection{Deep Learning Method}

In recent years, with the increase in the volume of datasets, the number and complexity of objects of interest have also grown. Additionally, due to the powerful feature extraction capabilities of deep learning networks, deep learning methods perform well on image CD tasks  \cite{he2015deep, dosovitskiy2020image, sun2019deep, ru2020multi, peng2019end, zheng2021change, zhan2017change, zhang2018triplet, zhou2023unified}. Deep learning methods mainly use the encoder-decoder architecture for change detection. The encoder extracts discriminative features from images by multiple convolutional layers. The common feature extraction networks are VGG\cite{daudt2018fully}, ResNet\cite{he2015deep}, MobileNet \cite{howard2017mobilenets, sandler2019mobilenetv2}. Moreover, due to the capability of attention mechanisms to capture long-range dependencies, structures like Vision Transformers (ViT), Swin Transformer, and others are increasingly being adopted\cite{vaswani2023attention, fang2021snunet, bandara2022transformer,zhang2023relation,li2022transunetcd,zhang2022swinsunet, niu2023smnet}. The decoder enhances image resolution and details through deconvolution layers or linear interpolation layers. It achieves multi-level feature fusion by employing methods such as addition, absolute difference, or attention-based mechanisms \cite{zhang2021escnet}. After that, we can treat change detection as a classification task by utilizing methods like cross-entropy loss to obtain a change map. Alternatively, we can approach change detection as a metric learning task, using contrastive loss to generate the change map. 

Existing deep learning methods mainly improve the change detection performances from the following aspects: (a) increasing the receptive field size of the encoder and capturing rich context information\cite{fang2021snunet, chen2021remote}, (b) enhancing the interaction between multi-temporal images \cite{fang2023changer, zhang2021escnet, fang2023changer}, (i.e., feature difference \cite{daudt2018fully}, feature merging, image concatenation), (c) improving feature fusion based on the attention module \cite{huang2024spatiotemporal, daudt2018fully, chen2020spatial, chen2020dasnet, chen2022fccdn, li2022densely}, (d) adapting the loss function to alleviate the issue of long-tail distribution present in the dataset \cite{fang2021snunet, zhan2017change}, and (e) leveraging general knowledge from large-scale base models such as SAM\cite{kirillov2023segment}, Remoteclip \cite{liu2024remoteclip}, and others in order to reduce the reliance of existing change detection models on a large amount of labeled data and further enhancing model performance \cite{li2024new}. 



 Although existing deep learning methods have achieved outstanding change detection performances, they should train a specific deep network for each dataset. The dataset-specific deep network usually performs poor generalization on other datasets. This paper designs a change adapter network to improve the generalization of CD, which can conduct CD well on various datasets at the same time. 

\section{Method}\label{sectionmethods}
\subsection{Problem formulation}
Suppose that the multi-temporal images are represented as $(X_{1}, X_{2}) \in R^{3 \times H\times W}$, image change detection aims to predict the corresponding pixel-wise binary change label, $\hat Y \in R^{H\times W}$.
\begin{equation}
\hat{Y}=M(X_{1}, X_{2}),\\
\end{equation}
where $\hat{Y} \in \{0, 1\}$ is the predicted binary change label and $M$ represents the deep learning module for change detection. 

The deep learning module can be optimized by the cross-entropy loss: 
\begin{equation}
L=-\frac{1}{N}\sum_{i=1}^{N}\sum_{j=1}^{HW}y_{ij}log(\hat{y_{ij}})\\
\end{equation}
where $y_{ij}$ and $\hat{y_{ij}}$ denote the ground truth and prediction results of the $j_{th}$ pixel and the $i_{th}$ input image pairs, respectively, $N$ is the number of multi-temporal image pairs.

\subsection{Network backbone} 
Following previous deep networks, this paper also uses the encoder-decoder network structure. The encoder mainly contains several convolutional blocks to extract discriminative features from images. The decoder performs multi-level feature fusion and increases the feature resolution by deconvolutional layers. 

The extracted discriminative features of multi-temporal images through the encoder can be represented as follows:
\begin{equation}
\begin{aligned}
&f_{i,1}=E(X_1),\\
&f_{i,2}=E(X_2), i=1\dots l_e,\\
\end{aligned}
\end{equation}
where $E$ means the encoder operator, $f_{i,1}$ and $f_{i,2}$ are extracted features of multi-temporal images in the $i_{th}$ layer of the encoder, $l_e$ means the number of feature layers in the encoder. 

The high-level features contain rich semantic information but lack spatial information, which will hinder accurate pixel-wise predictions. The low-level features have fine spatial information but weak semantic information. In the decoder, CANet designs several feature fusion blocks (FF) to aggregate the multi-level features for more accurate change label prediction. As shown in Fig. \ref{FRAMEWORK}, FF block contains a CBAM module \cite{woo2018cbam}, a deconvolutional layer, and a convolutional layer. Specifically, the first FF block aggregates the features of multi-temporal images, and other FF blocks merge the features of the multi-temporal images and the output of the previous FF block. The feature fusion process can be formulated as follows:
\begin{equation}
\begin{aligned}
&f'_1=FF(f_{l_e,1}, f_{l_e,2}),\\
&f'_i=FF(f'_{i-1}, f_{l_e-i+1,1}, f_{l_e-i+1,2}),\\
\end{aligned}
\end{equation}
where $FF$ represents the feature fusion operator, $f'_i$ is the $i_{th}$ fused multi-level features in the decoder, $i=1\dots l_d $, and $l_d$ means the number layers in the decoder. 

In this paper, we present the CD performances based on various network backbones, such as MobileNetV2 \cite{sandler2019mobilenetv2}, ResNet18 \cite{he2015deep}, IBN-Net18-a \cite{pan2020once}, and IBN-Net18-b \cite{pan2020once}.


Existing deep learning methods are merely suitable for a single dataset CD. They should train a dataset-specific deep network for each dataset. When we apply the trained deep network to other datasets, its change detection performances will significantly degrade. That is to say, the trained deep network based on historical datasets cannot be used for new datasets. A fully independent deep network should be trained from scratch for the new dataset CD, significantly increasing the parameter storage and training costs. 

A parameter-efficient and training-efficient deep network should have strong generalization performances for various datasets and can quickly adapt to new datasets CD. To achieve this, we propose a change adapter network (CANet) to improve the generalization of change detection. 
\subsection{CANet}
As shown in Fig. \ref{FRAMEWORK}, CANet contains a dataset-shared learning module ($M_s$) and a dataset-specific learning module, adapter ($M_a$).
\begin{equation}
\hat{Y}=M_a(M_s(X_{1}, X_{2})),\\
\end{equation}

$M_s$ is shared for different datasets, $M_a$ is a lightweight unique module for each dataset. We can employ the trained $M_s$ module based on the historical dataset and quickly update the $M_a$ for new datasets CD.

Considering the common feature learning global of change detection, the encoder, and the first several decoder parts are shared between different datasets. To acquire more generalized change detection, CANet should deal with the significant differences in data distribution and labeling between different datasets. CANet designs a dataset-specific module, adapter, for each dataset. Additionally, to quickly adapt to new datasets, the adapter should have a lightweight structure with a small number of network parameters. Thus, we use the last several FF blocks and a convolutional block in the decoder to build the adapter.

\subsection{Adapter}
Due to different datasets concerning different kinds of change objects, we design an interesting change region mask block (ICM) in the adapter. ICM can adaptively capture interesting objects in different datasets and boost change detection performances. As shown in the right of \cref{FRAMEWORK}, ICM generates the mask ($m$) through attention learning mechanisms and outputs the masked change prediction.
\begin{equation}
\hat{Y}= m \odot p,\\
\end{equation}
where $p$ is the output of the last convolutional block, $p\in R^{2 \times H\times W}$. 

Specifically, the ICM contains convolutional layers, maxpooling (MP), average pooling (AP), feature concatenation, and Squeeze-and-Excitation (SE) block \cite{hu2018squeeze}. Suppose the input of the ICM is $p \in R^{2\times H \times W}$. The details of the mask generator process can be formulated as follows:
\begin{equation}
\begin{aligned}
&x=conv(p),\\
&x_f=x \oplus MP(x) \oplus AP(x), \\
&x_m=conv(SE(x_f)),\\
&m= \delta(x_m) \oplus (1-\delta(x_m)),\\
\end{aligned}
\end{equation}
where $conv$ represents the convolutional operator, $\oplus$ is concatenation along the channel dimension, $\delta$ represents the sigmoid function.  

Additionally, different datasets have various data distributions caused by illumination changes, season changes, and sensor changes. Thus, CANet employs unique batch normalization (BN) layers for each dataset, which can tolerate image differences in appearance and style.

\begin{table*}[ht]
\centering
\caption{Details of the four public datasets for change detection.}
\begin{tabular}{@{}lcccccl@{}}
\hline
 \multirow{2}{*}{\textbf{Dataset}} &  \multirow{2}{*}{\textbf{Resolution(m)}} & \multirow{2}{*}{\textbf{Size}} & \multicolumn{3}{c}{\textbf{Number of Image Pairs}} & \multirow{2}{*}{\textbf{Interested change objects}} \\ \cline{4-6}
 &  &  & \textbf{Train} & \textbf{Test} & \textbf{Val} &  \\ \hline
CDD & 0.03-1 & 256$\times$256 & 10000 & 3000 & 3000 & Building, Road, Car, etc. \\
SYSU & 0.5 & 256$\times$256 & 12000 & 4000 & 4000 & Building, Road, Vegetation, etc. \\
LEVIR & 0.3 & 256$\times$256 & 7120 & 2048 & 1024 & Building \\
WHU & 0.2 & 256$\times$256 & 4571 & 1524 & 1525 & Building \\ \hline
\end{tabular}
\label{informationdataset}
\end{table*}

\subsection{Discussion}
Unlike the general domain adaption methods, which focus on improving the performance of new datasets (target dataset), CANet performs well on historical datasets (source dataset) and generalizes well on new datasets. First, we train CANet based on the historical dataset. Then, CANet shares the dataset-shared module with the new dataset and only optimizes the dataset-specific module (adapter) for the new dataset. In this way, CANet can detect changes in historical and new datasets at the same time. 

In conclusion, the advantages of the proposed CANet are as follows:
\begin{itemize}
    \item \textbf{Parameter efficiency.} Compared with existing deep networks that are merely trained for a specific dataset, our proposed CANet can be used for various datasets simultaneously. Thus, CANet has higher parameter efficiency than other deep learning networks.
    
    \item \textbf{Training efficiency.} Compared with training a fully independent network from scratch for new datasets, CANet can be quickly and adaptively applied to new datasets by updating the lightweight adapter. Thus, CANet has a distinct advantage in terms of training efficiency. 
\end{itemize}

\section{Experiments}
\label{exper}
\subsection{Datasets and training details}
\textbf{Datasets.} We test the performance of our proposed CANet on four public change detection datasets: CDD \cite{lebedev2018change}, SYSU \cite{shi2021deeply}, LEVIR \cite{chen2020spatial}, and WHU \cite{ji2018fully}. The image resolution, image size, numbers, and interested change objects are shown in Table \ref{informationdataset}.

\noindent \textbf{Implemental Details.} We use the random horizontal flip for data augmentation. The optimizer is SGD, weight decay is $1e^{-4}$, the momentum is  0.9, batch size is 8, and epoch is 300. We present the performances of the CANet based on various network backbones, such as MobileNetV2 \cite{sandler2019mobilenetv2}, ResNet18 \cite{he2015deep}, IBN-Net18-a \cite{pan2020once}, and IBN-Net18-b \cite{pan2020once}, which are denoted as CANet[M], CANet[R], CANet[Ia], and CANet[Ib], respectively. CANet-O represents training a dataset-specific CANet from scratch for each dataset. Unless otherwise specified, CANet is trained on the CDD dataset (historical dataset) and then optimizes a unique adapter for each new dataset.

\noindent \textbf{Evaluation Protocols.} This paper uses F1-score (F1), Precision (P), Recall (R), and Intersection over Union (IoU) to evaluate the change detection performances. The higher these metrics are, the better the change detection performances are. 
\begin{table}[tbp]
\footnotesize
\caption{Change detection results of different models on the CDD dataset. [M], [R], [Ia], and [Ib] indicate that CANet uses the network backbone of MobileNetV2, ResNet18, IBN-Net18-a, and IBN-Net18-b, respectively. CANet-O means the fully dataset-specific deep network, which is trained from scratch for each dataset.}
\begin{center}
\begin{tabular}{@{}lcccc@{}}
\toprule
\multirow{2}*{Method} & \multicolumn{4}{c}{CDD}\\
\cline{2-5} & F1(\%) & P(\%) & R(\%) & IoU(\%) \\
\midrule
FC-EF        & 72.31 & 66.69 & 78.97 & 56.63 \\
FC-Siam-diff & 80.69 & 70.56 & 94.21 & 67.62 \\
FC-Siam-conc & 84.20 & 81.02 & 87.64 & 72.71\\
BiDateNet   & 91.97 & 89.12 & 95.00 & 85.13\\
STANet      & 90.81 & 88.36 & 93.39 & 83.16\\
DSAMNet     & 91.32 & 90.58 & 92.06 & 84.02\\
SNUNet      & 96.82 & 96.86 & 96.78 & 93.84\\
BAN         & 96.80 & 97.51 & 96.11 & 93.81\\
ChangeClip  & 97.85 & 97.88 & 97.81 & 95.78 \\
\hline
CANet-O[M]  & 97.84 & 97.66 & 98.03 & 95.78 \\
CANet-O[R]  & \textbf{97.95} & \textbf{97.93} & 97.98 & \textbf{95.99}\\
CANet-O[Ia] & 97.82 & 97.52 & \textbf{98.12} & 95.73\\
CANet-O[Ib] & 97.66 & 97.49 & 97.82 & 95.42 \\
\bottomrule
\end{tabular}
\label{CDD_all}
\end{center}
\end{table}

\begin{table*}[tbp]
\begin{center}
\footnotesize
\caption{Change detection results of different models on SYSU, LEVIR, and WHU datasets. [M], [R], [Ia], and [Ib] indicate that CANet uses the network backbone of MobileNetV2, ResNet18, IBN-Net18-a, and IBN-Net18-b, respectively. CANet-O means the fully dataset-specific deep network for each dataset. Param indicates the number of parameters involved in the update during model training.}
\begin{tabular}{@{}lccccccccccccccc@{}}
\toprule
\multirow{2}*{Method}&\multicolumn{4}{c}{SYSU} && \multicolumn{4}{c}{LEVIR} && \multicolumn{4}{c}{WHU} & 
 Param\\
\cline{2-5} \cline{7-10}\cline{12-15} & F1(\%) & P(\%) & R(\%) & IoU(\%) & &F1(\%) & P(\%) & R(\%) & IoU(\%)&&F1(\%) & P(\%) & R(\%) & IoU(\%)& (M)\\
\midrule
\multicolumn{14}{c}{\textbf{Dataset-specific networks}}\\
FC-EF           & 75.07 & 74.32 & 75.84 & 60.09 && 80.74 & 79.41 & 82.12 & 67.70 && 75.92 & 79.20 & 72.90 & 61.19 & 1.35  \\
FC-Siam-diff   & 72.57 & \textbf{89.13} & 61.21 & 56.96 && 87.47 & 83.17 & 92.22 & 77.72 && 73.50 & 84.05 & 65.31 & 58.10 & 1.35  \\ 
FC-Siam-conc    & 77.21 & 73.35 & 81.48 & 62.87 && 85.56 & 79.99 & 92.02 & 74.81 && 80.81 & 74.18 & 88.75 & 67.80 & 1.55  \\ 
BiDateNet       & 78.58 & 80.30 & 76.92 & 64.71 && 89.32 & 86.68 & 92.12 & 80.70 && 85.70 & 80.36 & 91.81 & 74.99 & 13.40 \\ 
STANet          & 77.19 & 73.99 & 80.28 & 62.86 && 87.32 & 86.71 & 87.95 & 77.50 && 86.98 & 90.92 & 83.37 & 76.96 & 12.21 \\ 
DSAMNet         & 75.11 & 81.67 & 69.52 & 60.14 && 84.36 & 86.34 & 82.46 & 72.94 && 80.01 & 83.47 & 76.84 & 66.69 & 16.95 \\
SNUNet          & 78.68 & 76.49 & 80.99 & 64.85 && 90.03 & 89.01 & 91.07 & 81.86 && 84.12 & 84.26 & 83.98 & 72.60 & 27.07 \\
BAN             & 81.07 & 89.10 & 74.37 & 68.17 && 90.70 & \textbf{92.51} & 88.96 & 82.99 && 92.17 & \textbf{96.91} & 87.87 & 85.48 & 3.8 \\
ChangeClip      & 83.32 & 87.16 & 79.80 & 71.41 && 90.91 & 92.42 & 89.44 & 83.33 && 92.25 & 95.07 & 89.60 & 85.62 & - \\
\hline
\multicolumn{15}{c}{\textbf{Parameter-efficient deep networks}}\\
CANet-O[M]      & \textbf{83.91} & 82.78 & \textbf{85.07} & \textbf{72.27} && \textbf{91.17} & 90.30 & 92.07 & \textbf{83.78} && \textbf{92.83} & 91.42 & 94.28 & \textbf{86.62} & 10.48 \\
CANet[M]        & 81.57 & 80.94 & 82.21 & 68.88 && 89.83 & 89.35 & 90.32 & 81.54 && 91.17 & 88.32 & 94.21 & 83.77 & \textbf{0.43}  \\
CANet-O[R]      & 82.91 & 81.41 & 84.47 & 70.81 && 90.80 & 89.39 & \textbf{92.25} & 83.14 && 91.82 & 89.02 & 94.80 & 84.88 & 20.39 \\
CANet[R]        & 81.63 & 79.30 & 84.11 & 68.97 && 90.06 & 88.43 & 91.74 & 81.91 && 90.51 & 88.12 & 93.03 & 82.67 & 1.57 \\
CANet-O[Ia]     & 83.41 & 82.51 & 84.33 & 71.54 && 90.88 & 90.00 & 91.78 & 83.29 && 91.63 & 89.22 & 94.17 & 84.55 & 22.53 \\
CANet[Ia]       & 80.68 & 78.38 & 83.11 & 67.61 && 89.27 & 87.46 & 91.15 & 80.62 && 90.24 & 88.25 & 92.31 & 82.21 & 1.57  \\
CANet-O[Ib]     & 83.79 & 83.04 & 84.55 & 72.10 && 90.73 & 89.43 & 92.07 & 83.04 && 92.27 & 89.76 & \textbf{94.92} & 85.64 & 22.53 \\
CANet[Ib]       & 80.37 & 79.76 & 80.99 & 67.18 && 88.90 & 87.81 & 90.03 & 80.02 && 89.93 & 88.23 & 91.69 & 81.70 & 1.57 \\
\bottomrule
\end{tabular}
\label{SYSU_LEVIR_WHU_all}
\end{center}
\vspace{-3mm}
\end{table*}


\begin{figure*}
    \centering
    \subfloat[SYSU]{\includegraphics[width=0.30\textwidth]{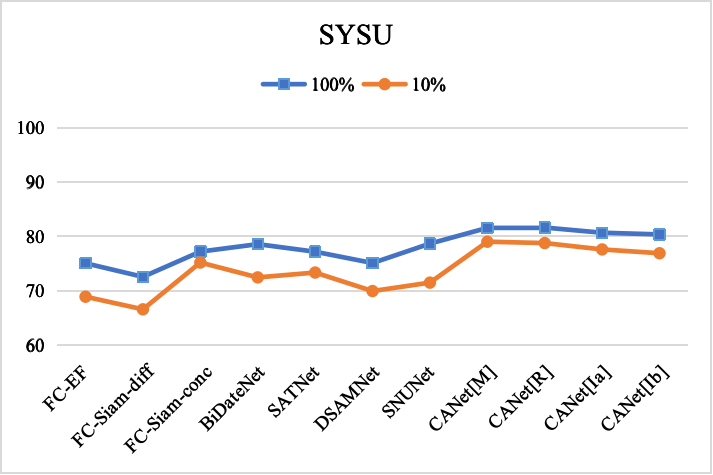}}
    \hfill
    \subfloat[LEVIR]{\includegraphics[width=0.30\textwidth]{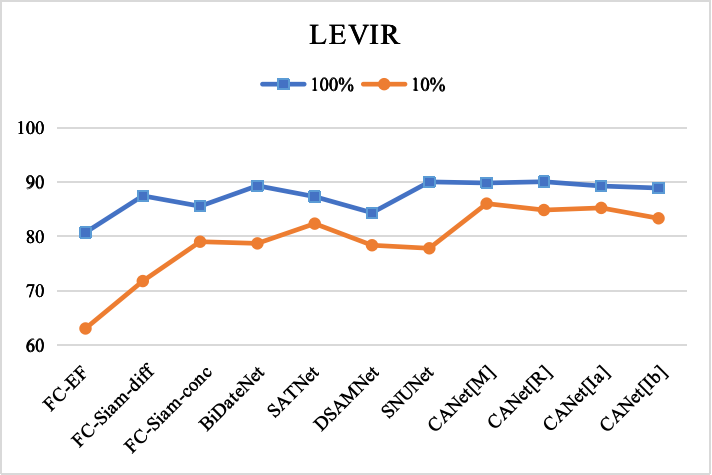}}
    \hfill
    \subfloat[WHU]{\includegraphics[width=0.30\textwidth]{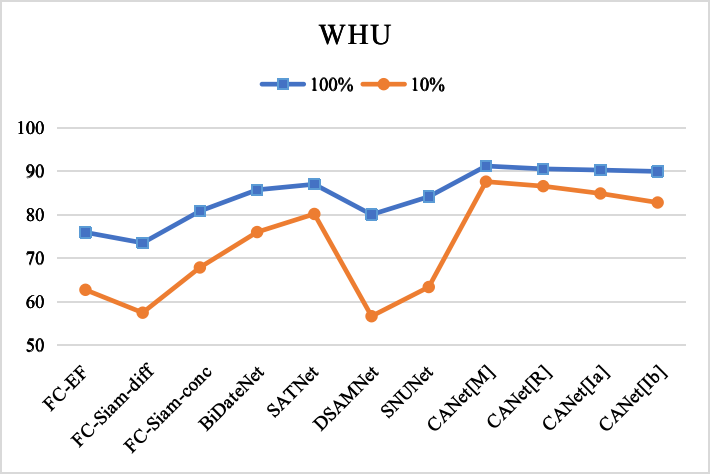}}
    \hfill
    \caption{F1 results of SYSU, LEVIR, and WHU datasets based on different numbers of training samples, i.e., 10\% and 100\%.}
    \label{few}
\end{figure*}


\begin{figure*}[htbp]
    \centering
    \subfloat[SYSU]{\includegraphics[width=0.30\textwidth]{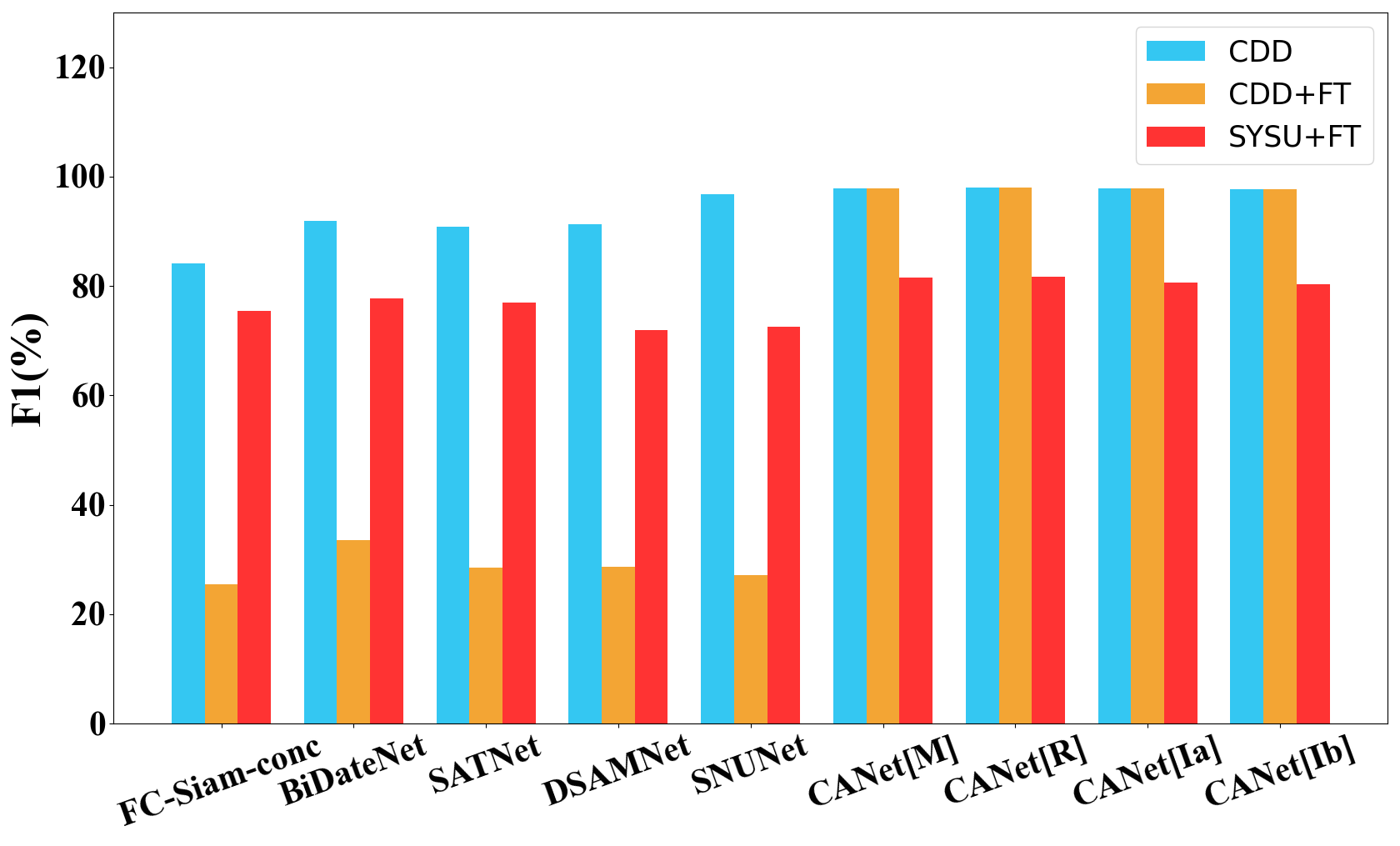}}
    \hfill
    \subfloat[LEVIR]{\includegraphics[width=0.30\textwidth]{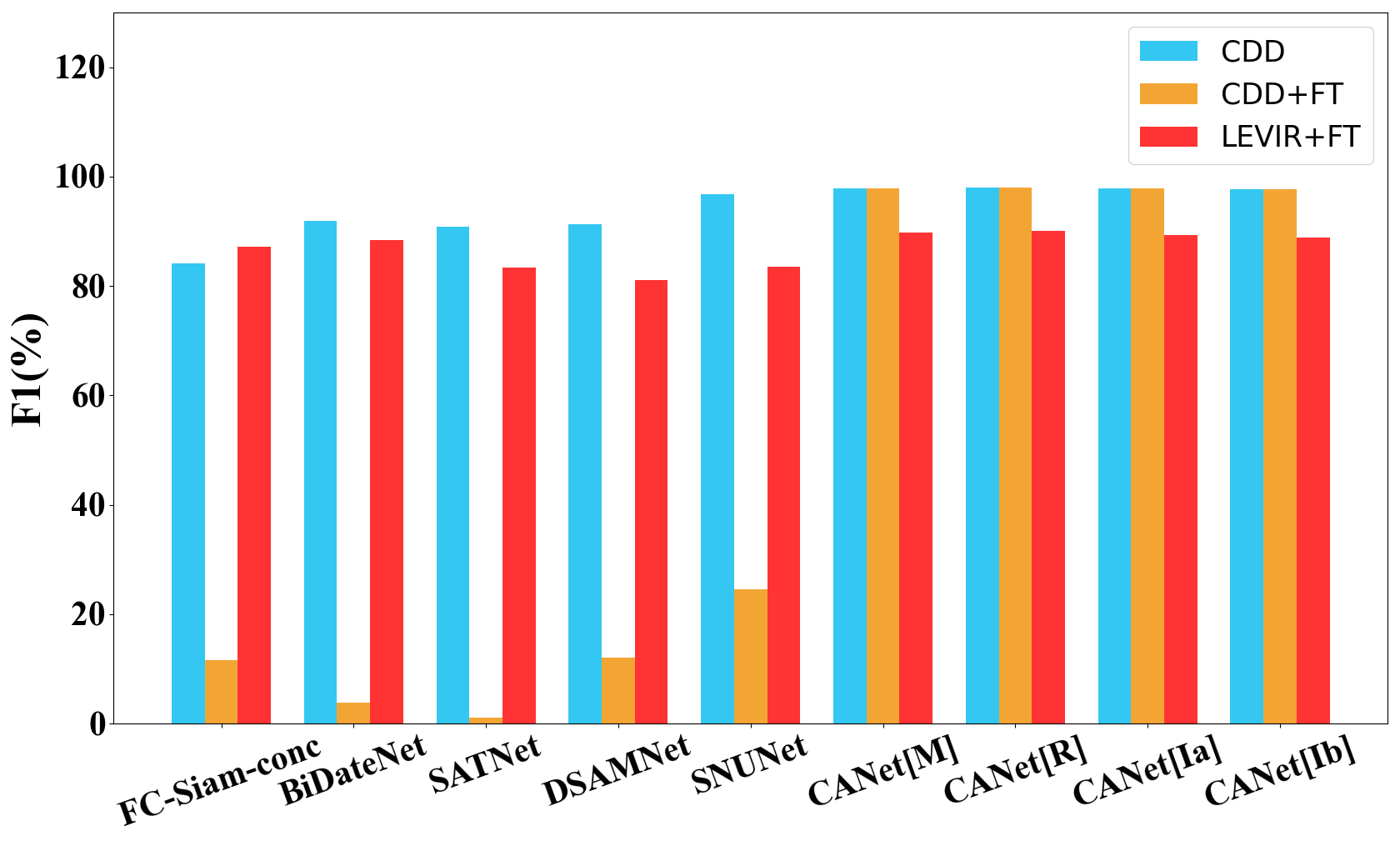}}
    \hfill
    \subfloat[WHU]{\includegraphics[width=0.30\textwidth]{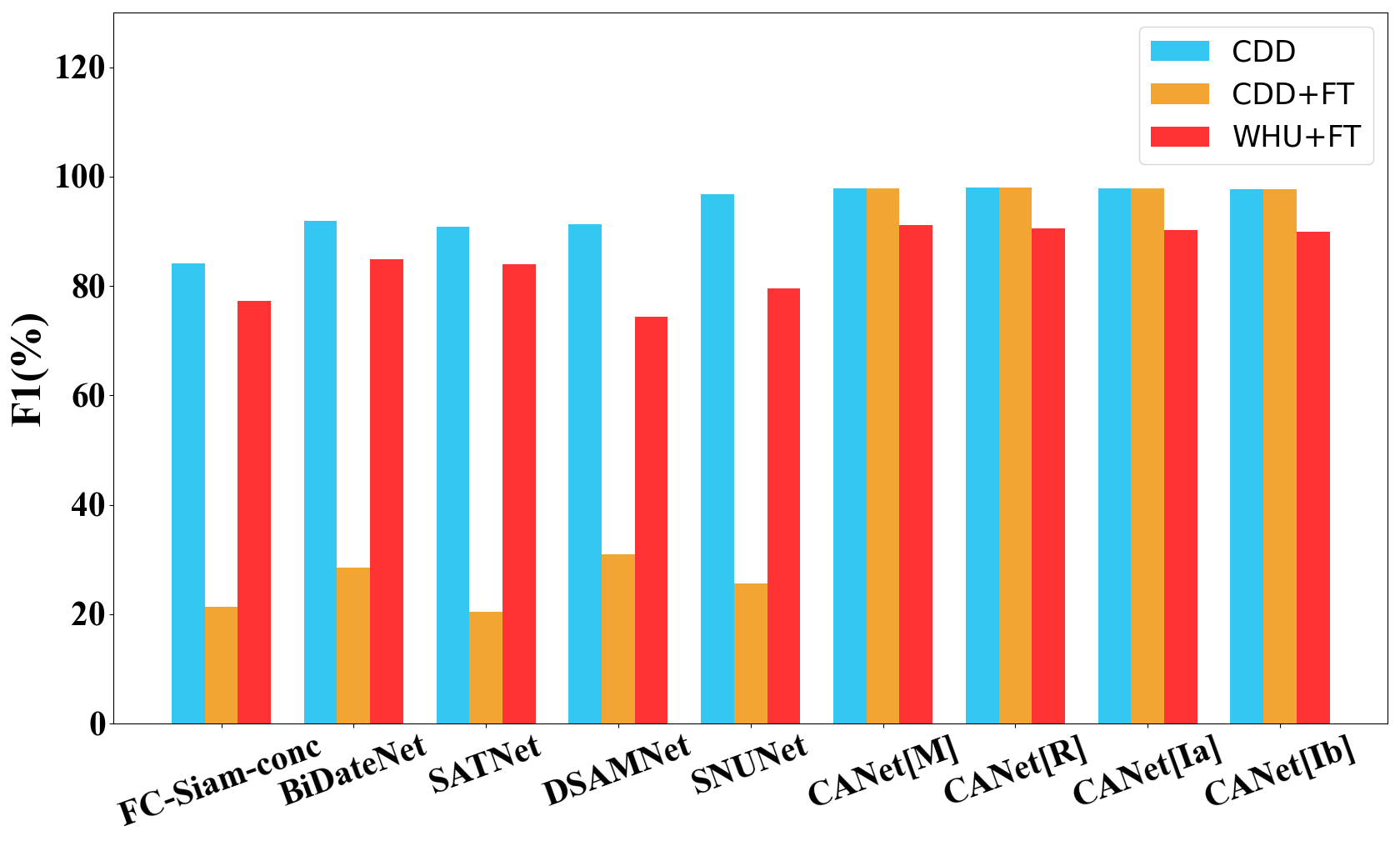}}
    \hfill
    \caption{The results of the original deep models on the historical dataset (CDD) and the results of the updated deep models through the online training strategy on the historical (CDD+FT) and new datasets (SYSU+FT, LEVIR+FT, and WHU+FT). FT means the deep model is trained on the CDD dataset and then fine-tuned on the new dataset.}
    \label{oline}
\end{figure*}

\begin{table*}[tbp]
\begin{center}
\footnotesize
\caption{Ablation experimental results based on different Adapter settings. CANet uses MobileNetV2 as the network backbone.}
\begin{tabular}{@{}lcccccccccccc@{}}
\toprule
\multirow{-3}{*}{}& \multicolumn{4}{c}{SYSU (\%)} & \multicolumn{4}{c}{LEVIR (\%)} & \multicolumn{4}{c}{WHU (\%)}   \\
\cline{2-13} & F1         & P          & R          & IoU       & F1   & P    & R    & IoU  & F1   & P    & R    & IoU  \\
\midrule
W/O ICM         & 79.28 & 78.07 & 80.52 & 65.67 & 89.13 & 87.06 & 91.30 & 80.39 & 87.98 & 84.75 & 91.46 & 78.53 \\
W/S-ICM       & 80.86 & 79.19 & \textbf{82.62} & 67.88 & 89.75 & 88.02 & \textbf{91.54} & 81.40 & 90.52 & 87.28 & 94.00 & 82.68 \\
W/O BN       & 79.57 & 79.93 & 79.21 & 66.07 & 88.59 & 86.31 & 91.00 & 79.52 & 84.32 & 87.78 & 81.12 & 72.89 \\
CANet(\textbf{ours})        & \textbf{81.57} & \textbf{80.94} & 82.21 & \textbf{68.88} & \textbf{89.83} & \textbf{89.35} & 90.32 & \textbf{81.54} & \textbf{91.17} & \textbf{88.32} & \textbf{94.21} & \textbf{83.77} \\
\bottomrule
\end{tabular}
\label{ablateone}
\end{center}
\vspace{-3mm}
\end{table*}

\begin{table*}[tbp]
\footnotesize
\caption{The impact of parameter $\eta$ in the adapter module. Param indicates the number of parameters involved in the training. In this experiment, CANet uses MobileNetV2 as the network backbone.}
\begin{center}
\begin{tabular}{@{}lccccccccccccc@{}}
\toprule
\multirow{3}*{  $\eta$  } & \multicolumn{4}{c}{SYSU (\%)} & \multicolumn{4}{c}{LEVIR (\%)} & \multicolumn{4}{c}{WHU (\%)} & \multirow{3}*{Param(M)}\\
\cline{2-13} & F1         & P          & R          & IoU       & F1   & P    & R    & IoU  & F1   & P    & R    & IoU  \\
\midrule
5   & \textbf{81.57} & 80.94 & 82.21 & \textbf{68.88} & 89.83 & \textbf{89.35} & 90.32 & 81.54 & \textbf{91.17} & \textbf{88.32} & 94.21 & \textbf{83.77} & 0.43  \\
4   & 81.15 & 79.34 & \textbf{83.05} & 68.28 & \textbf{89.91} & 88.13 & \textbf{91.76} & \textbf{81.67} & 90.80 & 86.95 & 95.01 & 83.16 & 0.15  \\
3   & 80.86 & \textbf{81.11} & 80.61 & 67.86 & 88.82 & 86.91 & 90.82 & 79.89 & 90.85 & 86.42 & \textbf{95.75} & 83.23 & 0.07  \\
2   & 79.57 & 77.64 & 81.58 & 66.07 & 88.07 & 86.25 & 89.96 & 78.68 & 89.95 & 86.48 & 93.72 & 81.74 & \textbf{0.04}  \\
\bottomrule
\end{tabular}
\label{ablatetwo}
\end{center}
\end{table*}
\subsection{Comparative experiment}
Table \ref{CDD_all} and Table \ref{SYSU_LEVIR_WHU_all} compare change detection performances and the number of network parameters of the proposed method with existing deep learning methods on CDD, SYSU, LEVIR, and WHU datasets. From these experimental results, we can draw the following conclusions. 

 Firstly, CANet-O achieves the best performances on four public datasets, as shown in Table \ref{CDD_all} and Table \ref{SYSU_LEVIR_WHU_all}. For example, the best F1 result of CAMNet-O is 97.95\% in the CDD, 83.91\% in the SYSU, 91.17\% in the LEVIR, and 92.83\% in the WHU, while the best F1 result of other dataset-specific networks is 96.82\% in the CDD, 76.68\% in the SYSU, 90.03\% in the LEVIR, 86.98\% in the WHU. Our proposal increases the F1 result of other deep networks by 1.13\% in the CDD, increases by 7.23\% in the SYSU, increases by 1.14\% in the LEVIR, and increases by 5.85\% in the WHU. The average improvement of F1 result brought by our CANet on four test datasets is 3.84\%. These experimental results demonstrate the network structure advantages of our CANet.

Secondly, CANet presents a slight performance decrease compared to the fully dataset-specific network of CANet-O. However, the significant advantage of our method is that CANet has a powerful generalization capability, which can conduct change detection well on various datasets simultaneously. CANet-O and other deep learning methods are merely suitable for change detection on a single dataset. For example, the average difference of F1 value on three new datasets (SYSU, LEVIR, and WHU) between CANet[M] and CANet-O [M] is 1.78\%, but the training network parameters is decreased by 95.90\% for each new dataset. The average difference of F1 value on three new datasets between CANet[R] and CANet-O [R] is 1.11\%, but the training network parameters are decreased by 92.30\% for each new dataset.


Thirdly, CANet achieves good change detection results based on different network backbones. It indicates the high flexibility and universality of our proposed CANet, which can combined with existing deep networks for more generalized change detection. 

Finally, the lightweight adapter significantly increases the parameter and training efficiency. The number of parameters of the adapter is about 4.1\%-7.7\% of the whole network parameters in the CANet. We only train a small number of parameters to adapt to new datasets. Thus, CANet can be quickly generalized to new datasets with small training costs. Additionally, different datasets share the most network parameters (92.3\%-95.9\%), greatly improving the parameter reuse probability and increasing parameter efficiency.

In summary, compared with existing deep learning networks, CANet has superior change detection performances, strong generalization ability, high flexibility, and small parameter memory and training costs on various datasets.

\subsection{Results on a few training datasets}
The CD performance of deep learning networks relies on a large number of labeled training datasets. However, it is difficult to acquire sufficient pixel-wise change labels for deep network optimization in practice. This part compares the CD results of various deep learning methods under a few training datasets, such as 10\% training datasets. As shown in Fig. \ref{few}, when the number of training datasets decreases from 100\% to 10\%, the CD performance of deep learning methods rapidly drops on SYSU, LEVIR, and WHU datasets, particularly FC-EF, BiDateNet, and STANet. However, there are very slight performance decreases in our CANet. On the one hand, the multi-level feature fusion module and ICM block can improve the CD performance by emphasizing the useful features. On the other hand, only the lightweight adapter needs to be optimized for new datasets. Thus, CANet reduces the requirement for many labeled training datasets, decreases the risk of overfitting, and achieves outstanding results under limited training datasets. These experimental results indicate that CANet has strong generalization and robustness, which can be generalized to various datasets, and performs detection well with few available training samples.


\subsection{Online training}
Online training is another feasible approach for generalizing to new datasets. The trained deep network based on the historical dataset (CDD) can be fine-tuned using new datasets. Figure \ref{oline} shows the results of the original deep models on the historical dataset and displays the results of the updated deep models on historical and new datasets. 

Firstly, online training can also achieve good change detection results on the new dataset. However, the performance of the updated deep models will degrade significantly on the old dataset. Our method not only performs well on the new datasets but also maintains better performance on the historical dataset. Secondly, our approach outperforms most deep models on both old and new datasets. Furthermore, CANet merely updates a small number of parameters in the adapter module for new datasets, which can be quickly expanded and applied to new datasets, significantly decreasing the training cost. These experimental results also prove the effectiveness and advantages of our CANet, which has high training efficiency and strong generalization on historical and new datasets.



\subsection{Ablation experiment}
This part verifies the effectiveness of the ICM block, unique BN layers, and the influences of the number of adapter layers. 

\textbf{ICM block.} Table \ref{ablateone} presents the results of CANet without the ICM block (W/O ICM), CANet with shared ICM block for various datasets (W/S-ICM), and the CANet with unshared ICM block (CANet). It can be seen that without the ICM block, the change detection performances of CANet will decrease. Additionally, CANet with the unshared ICM block acquires better results than CANet with the shared ICM block. The main reason is that the ICM block can enforce the deep model to adaptively emphasize the interested change objects in various datasets, which can reduce the influences of the significant differences in labeling. These results demonstrated the effectiveness of the proposed ICM block in the adapter module. 

\textbf{BN layers.} This part tests the influence of the unique BN layers. Various datasets have significant differences in image appearance. Thus, we design unique BN layers for each dataset. Table \ref{ablateone} compares the results of the adapter module with (CANet) and without unique BN layers (W/O BN) for each dataset. W/O BN means CANet shares BN layers in the dataset-shared learning module between various datasets. As shown in \cref{ablateone}, CANet with unique BN layers performs better than without BN layers. Specifically, compared with CANet, the F1 result of W/O BN decreases by 2.0\% on the SYSU dataset, decreases by 1.24\% on the LEVIR dataset, and decreases by 6.85\% on the WHU dataset. The main reason is that various datasets have significant differences in image styles caused by sensor and scene changes, resulting in different data distributions. Thus, CANet employs unshared BN layers for various datasets.

\textbf{Adapter layers.} CANet selects the latter $\eta$ blocks into the adapter. The large $\eta$ means more unshared parameters for various datasets, and the small $\eta$ means more shared parameters between different datasets and few dataset-specific network parameters for each dataset. As shown in Table \ref{ablatetwo}, when the $\eta$ decreases, the number of parameters involved in training for new datasets becomes smaller, and the network change detection performances have a slight degradation. The main reason is that there are a small number of network parameters for adapting the significant differences in image appearance and labeling. However, the CANet still acquires comparative performances on different datasets simultaneously and exceeds most dataset-specific deep networks. Considering the number of parameters in the adapter and the change detection performances on new datasets, we set $\eta=5$ in experiments. We can flexibly select the number of adapter layers in applications based on the performance requirement and device limitations.

\section{Conclusion}\label{sectionconclusion}
In this paper, we propose a change adapter network, CANet, with strong generalization performances for change detection. CANet contains a dataset-shared learning module for discriminative feature learning and a lightweight dataset-specific learning module for adapting the significant differences in data distribution and labeling. Unlike existing deep learning methods merely applied for the single dataset, CANet can simultaneously achieve better and more competitive change detection performances on various datasets. Additionally, it can quickly expand to new datasets by updating the lightweight adapter module, increasing the parameter and training efficiency. CANet has high flexibility, which can easily inserted into existing deep learning modules for change detection. It also has superior change detection performance under limited training samples. 
{
    \small
    \bibliographystyle{ieeenat_fullname}
    \bibliography{main}
}

\end{document}


\maketitle

In this supplementary material, we will further provide the effects of using different datasets as pre-training data for CANet, as well as the visualization comparison results between our method and other methods.

\section{Start from different datasets}
This part tests the results of the CANet training starting from other datasets, such as generalize from SYSU to CDD, from LEVIR to CDD, from CDD to WHU, from SYSU to WHU, and from LEVIR to WHU. As shown in Table. \ref{differntstarts}, CANet always performs well on the new dataset, although it is pre-trained on different historical datasets. Specifically, when the training starting datasets are CDD, SYSU, and LEVIR, the F1 result on the WHU is all about 91\%. These experimental results have demonstrated the strong robustness of our proposal. The network performances on the new dataset are almost unaffected by the starting point.

\section{Visualization results}
This section presents a visualization of the change detection outcomes achieved by various deep learning methodologies when applied to the SYSU, LEVIR, and WHU datasets. As depicted in Fig. \ref{vis}, the datasets exhibit substantial variability in data distribution, which poses a significant challenge to the generalizability of deep networks across different datasets. Furthermore, discrepancies in labeling arise due to the diverse nature of objects that are considered interesting changes. Our proposed CANet and CANet-O models demonstrate an exceptional ability to accurately pinpoint regions undergoing change. The change detection results produced by these models align closely with the ground truth. Notably, the boundaries of the detected change regions are more precisely defined in our approach. In contrast, alternative methods exhibit numerous inaccuracies, such as misclassified static areas and indistinct boundary edges, as evidenced by the results from DSAMNet and STANet on the SYSU dataset and those from DSAMNet and STANet on the LEVIR dataset.

\begin{table}[!tbp]
\footnotesize
\setlength{\tabcolsep}{6pt}
\caption{Change detection results of CANet based on different start training settings. SYSU$\to$  CDD means CANet is first trained on the SYSU and then adapter for the CDD.}
\begin{center}
\begin{tabular}{@{}lccccc@{}}
\toprule
Method&settings&F1(\%) & P(\%) & R(\%) & IoU(\%)\\
\midrule
CANet [M] &SYSU$\to$  CDD& 94.76 & 93.98 & 95.55 & 90.03 \\
CANet [M]&LEVIR$\to$  CDD& 93.39 & 91.95 & 94.88 & 87.60 \\
\midrule
CANet [M]& CDD$\to$  WHU&91.17 & 88.32 & 94.21 & 83.77\\
CANet [M]& SYSU$\to$  WHU&91.83 & 88.91 & 94.95 & 84.90 \\
CANet [M] &LEVIR$\to$  WHU&91.95 & 88.94 & 95.17 & 85.10 \\
\bottomrule
\end{tabular}
\label{differntstarts}
\end{center}
\end{table}

The comparison of these visual representations highlights the effectiveness of our proposed CANet and CANet-O models in accurately detecting changes within remote sensing imagery. The ability of our models to outperform competing methods in these key aspects underscores their potential for enhancing the reliability and precision of change detection applications in diverse real-world scenarios.

\begin{figure*}[!t]
\centering

\begin{subfigure}[t]{0.12\linewidth}
\centering
\includegraphics[width=2.0cm]{experiment/vis/SYSU/01019_A.png}
\end{subfigure}%
\begin{subfigure}[t]{0.12\linewidth}
\centering
\includegraphics[width=2.0cm]{experiment/vis/SYSU/01019_B.png}
\end{subfigure}%
\begin{subfigure}[t]{0.12\linewidth}
\centering
\includegraphics[width=2.0cm]{experiment/vis/SYSU/01019_L_F1.png}
\end{subfigure}%
\begin{subfigure}[t]{0.12\linewidth}
\centering
\includegraphics[width=2.0cm]{experiment/vis/SYSU/01019_DSAMNet_F1.png}
\end{subfigure}%
\begin{subfigure}[t]{0.12\linewidth}
\centering
\includegraphics[width=2.0cm]{experiment/vis/SYSU/01019_STANet_F1.png}
\end{subfigure}%
\begin{subfigure}[t]{0.12\linewidth}
\centering
\includegraphics[width=2.0cm]{experiment/vis/SYSU/01019_CANet_F1.png}
\end{subfigure}%
\begin{subfigure}[t]{0.12\linewidth}
\centering
\includegraphics[width=2.0cm]{experiment/vis/SYSU/01019_CANet_O_F1.png}
\end{subfigure}%
\vspace{5pt}

\begin{subfigure}[t]{0.12\linewidth}
\centering
\includegraphics[width=2.0cm]{experiment/vis/SYSU/00516_A.png}
\end{subfigure}%
\begin{subfigure}[t]{0.12\linewidth}
\centering
\includegraphics[width=2.0cm]{experiment/vis/SYSU/00516_B.png}
\end{subfigure}%
\begin{subfigure}[t]{0.12\linewidth}
\centering
\includegraphics[width=2.0cm]{experiment/vis/SYSU/00516_L.png}
\end{subfigure}%
\begin{subfigure}[t]{0.12\linewidth}
\centering
\includegraphics[width=2.0cm]{experiment/vis/SYSU/00516_DSAMNet_F1.png}
\end{subfigure}%
\begin{subfigure}[t]{0.12\linewidth}
\centering
\includegraphics[width=2.0cm]{experiment/vis/SYSU/00516_STANet_F1.png}
\end{subfigure}%
\begin{subfigure}[t]{0.12\linewidth}
\centering
\includegraphics[width=2.0cm]{experiment/vis/SYSU/00516_CANet_F1.png}
\end{subfigure}%
\begin{subfigure}[t]{0.12\linewidth}
\centering
\includegraphics[width=2.0cm]{experiment/vis/SYSU/00516_CANet_O_F1.png}
\end{subfigure}%
\vspace{5pt}

\begin{subfigure}[t]{0.12\linewidth}
\centering
\includegraphics[width=2.0cm]{experiment/vis/SYSU/03952_A.png}
\end{subfigure}%
\begin{subfigure}[t]{0.12\linewidth}
\centering
\includegraphics[width=2.0cm]{experiment/vis/SYSU/03952_B.png}
\end{subfigure}%
\begin{subfigure}[t]{0.12\linewidth}
\centering
\includegraphics[width=2.0cm]{experiment/vis/SYSU/03952_L.png}
\end{subfigure}%
\begin{subfigure}[t]{0.12\linewidth}
\centering
\includegraphics[width=2.0cm]{experiment/vis/SYSU/03952_DSAMNet_F1.png}
\end{subfigure}%
\begin{subfigure}[t]{0.12\linewidth}
\centering
\includegraphics[width=2.0cm]{experiment/vis/SYSU/03952_STANet_F1.png}
\end{subfigure}%
\begin{subfigure}[t]{0.12\linewidth}
\centering
\includegraphics[width=2.0cm]{experiment/vis/SYSU/03952_CANet_F1.png}
\end{subfigure}%
\begin{subfigure}[t]{0.12\linewidth}
\centering
\includegraphics[width=2.0cm]{experiment/vis/SYSU/03952_CANet_O_F1.png}
\end{subfigure}%
\vspace{5pt}

\begin{subfigure}[t]{0.12\linewidth}
\centering
\includegraphics[width=2.0cm]{experiment/vis/LEVIR/test_112_A.png}
\end{subfigure}%
\begin{subfigure}[t]{0.12\linewidth}
\centering
\includegraphics[width=2.0cm]{experiment/vis/LEVIR/test_112_B.png}
\end{subfigure}%
\begin{subfigure}[t]{0.12\linewidth}
\centering
\includegraphics[width=2.0cm]{experiment/vis/LEVIR/test_112_L_F1.png}
\end{subfigure}%
\begin{subfigure}[t]{0.12\linewidth}
\centering
\includegraphics[width=2.0cm]{experiment/vis/LEVIR/test_112_DSAMNet_F1.png}
\end{subfigure}%
\begin{subfigure}[t]{0.12\linewidth}
\centering
\includegraphics[width=2.0cm]{experiment/vis/LEVIR/test_112_STANet_F1.png}
\end{subfigure}%
\begin{subfigure}[t]{0.12\linewidth}
\centering
\includegraphics[width=2.0cm]{experiment/vis/LEVIR/test_112_CANet_F1.png}
\end{subfigure}%
\begin{subfigure}[t]{0.12\linewidth}
\centering
\includegraphics[width=2.0cm]{experiment/vis/LEVIR/test_112_CANet_O_F1.png}
\end{subfigure}%
\vspace{5pt}

\begin{subfigure}[t]{0.12\linewidth}
\centering
\includegraphics[width=2.0cm]{experiment/vis/LEVIR/test_735_A.png}
\end{subfigure}%
\begin{subfigure}[t]{0.12\linewidth}
\centering
\includegraphics[width=2.0cm]{experiment/vis/LEVIR/test_735_B.png}
\end{subfigure}%
\begin{subfigure}[t]{0.12\linewidth}
\centering
\includegraphics[width=2.0cm]{experiment/vis/LEVIR/test_735_L_F1.png}
\end{subfigure}%
\begin{subfigure}[t]{0.12\linewidth}
\centering
\includegraphics[width=2.0cm]{experiment/vis/LEVIR/test_735_DSAMNet_F1.png}
\end{subfigure}%
\begin{subfigure}[t]{0.12\linewidth}
\centering
\includegraphics[width=2.0cm]{experiment/vis/LEVIR/test_735_STANet_F1.png}
\end{subfigure}%
\begin{subfigure}[t]{0.12\linewidth}
\centering
\includegraphics[width=2.0cm]{experiment/vis/LEVIR/test_735_CANet_F1.png}
\end{subfigure}%
\begin{subfigure}[t]{0.12\linewidth}
\centering
\includegraphics[width=2.0cm]{experiment/vis/LEVIR/test_735_CANet_O_F1.png}
\end{subfigure}%
\vspace{5pt}

\begin{subfigure}[t]{0.12\linewidth}
\centering
\includegraphics[width=2.0cm]{experiment/vis/LEVIR/test_125_A.png}
\end{subfigure}%
\begin{subfigure}[t]{0.12\linewidth}
\centering
\includegraphics[width=2.0cm]{experiment/vis/LEVIR/test_125_B.png}
\end{subfigure}%
\begin{subfigure}[t]{0.12\linewidth}
\centering
\includegraphics[width=2.0cm]{experiment/vis/LEVIR/test_125_L_F1.png}
\end{subfigure}%
\begin{subfigure}[t]{0.12\linewidth}
\centering
\includegraphics[width=2.0cm]{experiment/vis/LEVIR/test_125_DSAMNet_F1.png}
\end{subfigure}%
\begin{subfigure}[t]{0.12\linewidth}
\centering
\includegraphics[width=2.0cm]{experiment/vis/LEVIR/test_125_STANet_F1.png}
\end{subfigure}%
\begin{subfigure}[t]{0.12\linewidth}
\centering
\includegraphics[width=2.0cm]{experiment/vis/LEVIR/test_125_CANet_F1.png}
\end{subfigure}%
\begin{subfigure}[t]{0.12\linewidth}
\centering
\includegraphics[width=2.0cm]{experiment/vis/LEVIR/test_125_CANet_O_F1.png}
\end{subfigure}%
\vspace{5pt}

\begin{subfigure}[t]{0.12\linewidth}
\centering
\includegraphics[width=2.0cm]{experiment/vis/WHU/14336_20736_A.png}
\end{subfigure}%
\begin{subfigure}[t]{0.12\linewidth}
\centering
\includegraphics[width=2.0cm]{experiment/vis/WHU/14336_20736_B.png}
\end{subfigure}%
\begin{subfigure}[t]{0.12\linewidth}
\centering
\includegraphics[width=2.0cm]{experiment/vis/WHU/14336_20736_L_F1.png}
\end{subfigure}%
\begin{subfigure}[t]{0.12\linewidth}
\centering
\includegraphics[width=2.0cm]{experiment/vis/WHU/14336_20736_DSAMNet_F1.png}
\end{subfigure}%
\begin{subfigure}[t]{0.12\linewidth}
\centering
\includegraphics[width=2.0cm]{experiment/vis/WHU/14336_20736_STANet_F1.png}
\end{subfigure}%
\begin{subfigure}[t]{0.12\linewidth}
\centering
\includegraphics[width=2.0cm]{experiment/vis/WHU/14336_20736_CANet_F1.png}
\end{subfigure}%
\begin{subfigure}[t]{0.12\linewidth}
\centering
\includegraphics[width=2.0cm]{experiment/vis/WHU/14336_20736_CANet_O_F1.png}
\end{subfigure}%
\vspace{5pt}

\begin{subfigure}[t]{0.12\linewidth}
\centering
\includegraphics[width=2.0cm]{experiment/vis/WHU/14592_23808_A.png}
\end{subfigure}%
\begin{subfigure}[t]{0.12\linewidth}
\centering
\includegraphics[width=2.0cm]{experiment/vis/WHU/14592_23808_B.png}
\end{subfigure}%
\begin{subfigure}[t]{0.12\linewidth}
\centering
\includegraphics[width=2.0cm]{experiment/vis/WHU/14592_23808_L_F1.png}
\end{subfigure}%
\begin{subfigure}[t]{0.12\linewidth}
\centering
\includegraphics[width=2.0cm]{experiment/vis/WHU/14592_23808_DSAMNet_F1.png}
\end{subfigure}%
\begin{subfigure}[t]{0.12\linewidth}
\centering
\includegraphics[width=2.0cm]{experiment/vis/WHU/14592_23808_STANet_F1.png}
\end{subfigure}%
\begin{subfigure}[t]{0.12\linewidth}
\centering
\includegraphics[width=2.0cm]{experiment/vis/WHU/14592_23808_CANet_F1.png}
\end{subfigure}%
\begin{subfigure}[t]{0.12\linewidth}
\centering
\includegraphics[width=2.0cm]{experiment/vis/WHU/14592_23808_CANet_O_F1.png}
\end{subfigure}%
\vspace{5pt}

\begin{subfigure}[t]{0.12\linewidth}
\centering
\includegraphics[width=2.0cm]{experiment/vis/WHU/1280_7936_A.png}
\caption{}
\end{subfigure}%
\begin{subfigure}[t]{0.12\linewidth}
\centering
\includegraphics[width=2.0cm]{experiment/vis/WHU/1280_7936_B.png}
\caption{}
\end{subfigure}%
\begin{subfigure}[t]{0.12\linewidth}
\centering
\includegraphics[width=2.0cm]{experiment/vis/WHU/1280_7936_L_F1.png}
\caption{}
\end{subfigure}%
\begin{subfigure}[t]{0.12\linewidth}
\centering
\includegraphics[width=2.0cm]{experiment/vis/WHU/1280_7936_DSAMNet_F1.png}
\caption{}
\end{subfigure}%
\begin{subfigure}[t]{0.12\linewidth}
\centering
\includegraphics[width=2.0cm]{experiment/vis/WHU/1280_7936_STANet_F1.png}
\caption{}
\end{subfigure}%
\begin{subfigure}[t]{0.12\linewidth}
\centering
\includegraphics[width=2.0cm]{experiment/vis/WHU/1280_7936_CANet_F1.png}
\caption{}
\end{subfigure}%
\begin{subfigure}[t]{0.12\linewidth}
\centering
\includegraphics[width=2.0cm]{experiment/vis/WHU/1280_7936_CANet_O_F1.png}
\caption{}
\end{subfigure}%
\centering

\caption{Visual comparison results on the SYSU (\emph{first three rows}), LEVIR (\emph{rows four to six}), and WHU (\emph{rows seven to nine})datasets. (a) and (b) are input multi-temporal images; (c) is the corresponding ground truth (GT); (d)-(g) are detection results of DSAMNet, STANet, CANet, and CANet-O, respectively.}
\label{vis}
\end{figure*}
